\documentclass{article}
\usepackage{spconf,amsmath,graphicx}

\usepackage{graphicx}
\usepackage[table,xcdraw]{xcolor}
\usepackage[normalem]{ulem}
\usepackage{color}
\usepackage{booktabs}
\useunder{\uline}{\ul}{}
\usepackage{amssymb}

\title{Learning Image Deraining Transformer Network with Dynamic Dual Self-Attention}
\name{Zhentao Fan \qquad Hongming Chen \qquad Yufeng Li}
\address{Gaofen Lab, Shenyang Aerospace University, Shenyang, China}
\begin{document}
%
\maketitle
\begin{abstract}
Recently, Transformer-based architecture has been introduced into single image deraining task due to its advantage in modeling non-local information.
However, existing approaches tend to integrate global features based on a dense self-attention strategy since it tend to uses all similarities of the tokens between the queries and keys.
In fact, this strategy leads to ignoring the most relevant information and inducing blurry effect by the irrelevant representations during the feature aggregation. 
To this end, this paper proposes an effective image deraining Transformer with dynamic dual self-attention (DDSA), which combines both dense and sparse attention strategies to better facilitate clear image reconstruction.
Specifically, we only select the most useful similarity values based on top-\emph{k} approximate calculation to achieve sparse attention.
In addition, we also develop a novel spatial-enhanced feed-forward network (SEFN) to further obtain a more accurate representation for achieving high-quality derained results.
Extensive experiments on benchmark datasets demonstrate the effectiveness of our proposed method.
\end{abstract}
\section{Introduction}
\label{sec:intro}
Single image deraining is a classical signal processing task emerging in the last decade, whose aim is to find a solution to the issue of recovering the clear and rain-free background from the observed rainy one \cite{yu2022towards,fan2023learning}. 
Compared with prior-based algorithms \cite{kang2011automatic}, \cite{li2016rain}, recent years have witnessed the success of deep learning in this field and numerous rain removal methods \cite{fu2017clearing}, \cite{chen2022unpaired}, \cite{yang2020single}, \cite{li2022tao} have been proposed.
Thanks to the powerful learning ability of convolutional neural networks (CNNs), the deep model have been investigated to memorize the correlation between the rainy image and its clear counterpart. These CNN-based networks have achieved decent restoration performance.

However, due to the local receptive fields and interactions of convolutions, CNNs have limited capability to model long-range dependency information \cite{zamir2022restormer}, \cite{xiao2022image}. 
To solve this bottleneck, recent researches have begun to introduce Transformers into various image restoration tasks \cite{dosovitskiy2020image,chen2023hybrid}.
As a new network backbone, Transformers have achieved significant improvements boost over CNN networks \cite{chen2022unpaired2,zamir2022restormer}, \cite{xiao2022image}, \cite{wang2022uformer}, which benefit from the natural advantage of self-attention with non-local feature aggregation. 
For example, Wang et al. \cite{wang2022uformer} proposes a general Transformer-based image restoration framework using U-Shaped architecture named Uformer. To reduce complexity, Restormer \cite{zamir2022restormer} introduces a multi-Dconv head transposed attention to capture local and non-local pixel interactions. For the field of image rain removal, Xiao et al. \cite{xiao2022image} elaborately develop image deraining Transformer (IDT) with window-based and spatial-based dual Transformer to obtain excellent results.

\begin{figure}[!t]
	\centering
	\includegraphics[width=1.0\columnwidth]{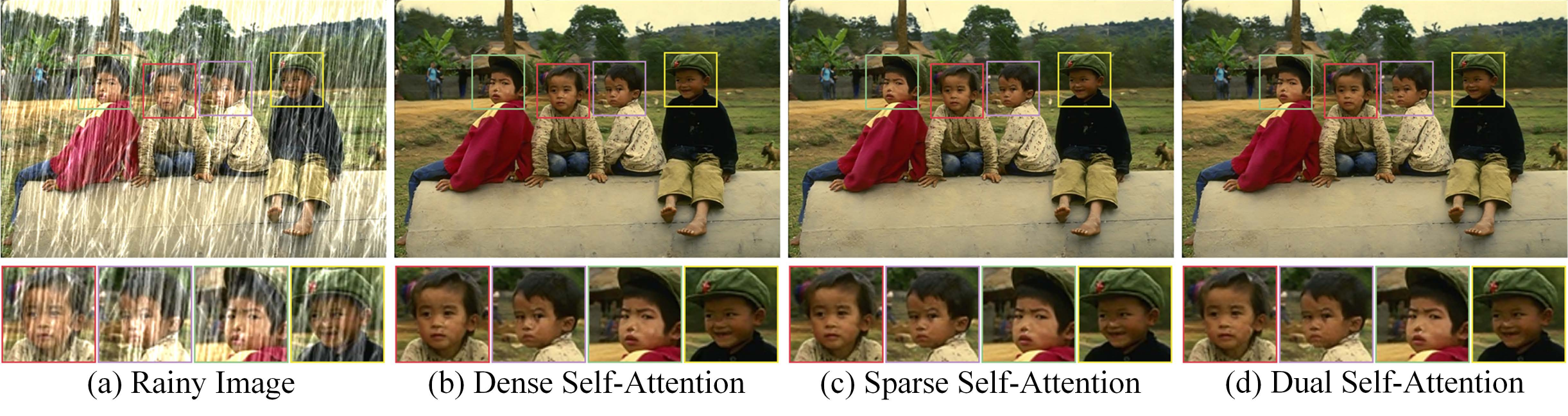} 
	\caption{Deraining results between our method and other self-attention based strategies. Our method can generate high-quality results with more accurate global textures and local details. Best viewed on high-resolution displays.} 
	\label{fig1}
\end{figure}

Albeit these methods have shown initial success, we observe that they may encounter practical difficulties when recovers the fine spatial details of images, especially in heavy rainy scenes highly degraded by intensive rain streaks.
In fact, these methods neglect a potential culprit that effected the quality of recovery results, \emph{i.e.}, the dense calculation pattern of self-attention in vanilla Transformer.
As we know, in the standard self-attention \cite{dosovitskiy2020image}, assigning all attention relations for the query-key pairs is the default operation.
Latest works \cite{zamir2022restormer}, \cite{xiao2022image}, \cite{wang2022uformer} attempt to directly apply this dense self-attention to Transformer-based image restorator.
This will naturally lead to their failure to focus the attention on the most relevant ones in the feature aggregation.
In other words, the integration of irrelevant features could cause blurry effects on the reconstructed results.
Therefore, this motivate us to explore more effective self-attention manner in the structure of Transformer.

\begin{figure*}[!t]
	\centering
	\includegraphics[width=1.0\textwidth]{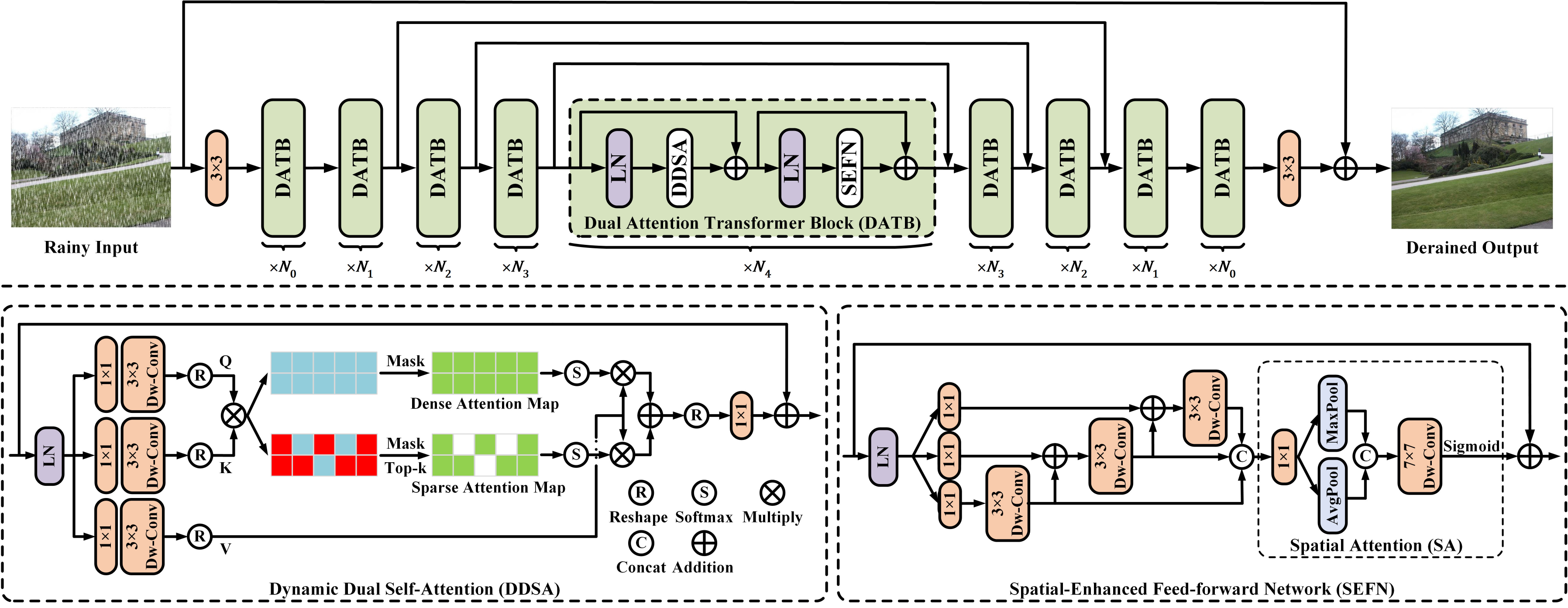} 
	\caption{The overall architecture of the proposed image deraining Transformer with dynamic dual self-attention for single image rain removal.}
	\label{fig2}
\end{figure*}

Towards this end, we propose an effective image deraining Transformer with dynamic dual self-attention (DDSA), which combines both dense and sparse attention strategies to better facilitate clear image reconstruction. 
Unlike the previous dense self-attention (DSA), we also select the most useful similarity values based on the top-\emph{k} approximate calculation to achieve sparse self-attention (SSA).
The combination of DSA and SSA can provide more flexible dynamic attention for the model to enrich both local and global representations.
In addition, we also develop an effective spatial-enhanced feed-forward network (SEFN) to further achieve spatial and channel integration simultaneously in Transformer. 
With above-mentioned designs, our method is capable of encouraging more effective feature aggregation and generating more accurate representation, in order to facilitate rain removal, see Fig. \ref{fig1}.

The contributions of this paper are summarized as follows: 

i) We design an effective dual self-attention in Transformer, which can dynamically and adaptively explore both local and global features for high-quality image reconstruction.

ii) We design a spatial-enhanced feed-forward network in Transformer, which can promote the integration of spatial-wise and channel-wise rain information in feature transformation. 

iii) Extensive experimental results on the commonly used benchmarks considerably demonstrate that our method outperforms existing state-of-the-art deraining approaches.
\section{Proposed Method}
\label{sec:format}
In this section, our proposed overall network architecture is presented in Fig. \ref{fig2}. Then, we describe two core components of the proposed network: dynamic dual self-attention (DDSA) and spatial-enhanced feed-forward network (SEFN).

\begin{table*}[htbp]
	\centering
	\caption{Comparison of quantitative results on five benchmark datasets. Bold and underline indicate the best and second-best results.}
	\resizebox{\textwidth}{!}{
		\begin{tabular}{ccccccccccccc}
			\hline
			Datasets      & \multicolumn{2}{c}{Test100 \cite{zhang2019image}} & \multicolumn{2}{c}{Rain100H \cite{yang2017deep}} & \multicolumn{2}{c}{Rain100L \cite{yang2017deep}} & \multicolumn{2}{c}{Test2800 \cite{fu2017removing}} & \multicolumn{2}{c}{Test1200 \cite{zhang2018density}} & \multicolumn{2}{c}{Average}     \\
			Methods         & PSNR           & SSIM           & PSNR            & SSIM           & PSNR            & SSIM           & PSNR            & SSIM           & PSNR            & SSIM           & PSNR           & SSIM           \\ \hline
			DerainNet \cite{fu2017clearing}  & 22.77          & 0.810          & 14.92           & 0.592          & 27.03           & 0.884          & 24.31           & 0.861          & 23.38           & 0.835          & 22.48          & 0.796          \\
			SEMI \cite{wei2019semi}       & 22.35          & 0.788          & 16.56           & 0.486          & 25.03           & 0.842          & 24.43           & 0.782          & 26.05           & 0.822          & 22.88          & 0.744          \\
			DIDMDN \cite{zhang2018density}     & 22.56          & 0.818          & 17.35           & 0.524          & 25.23           & 0.741          & 28.13           & 0.867          & 29.95           & 0.901          & 24.64          & 0.770          \\
			UMRL \cite{yasarla2019uncertainty}       & 24.41          & 0.829          & 26.01           & 0.832          & 29.18           & 0.923          & 29.97           & 0.905          & 30.55           & 0.910          & 28.02          & 0.880          \\
			RESCAN \cite{li2018recurrent}     & 25.00          & 0.835          & 26.36           & 0.786          & 29.80           & 0.881          & 31.29           & 0.904          & 30.51           & 0.882          & 28.59          & 0.858          \\
			PReNet \cite{ren2019progressive}     & 24.81          & 0.851          & 26.77           & 0.858          & 32.44           & 0.950          & 31.75           & 0.916          & 31.36           & 0.911          & 29.43          & 0.897          \\
			MSPFN \cite{jiang2020multi}      & 27.50          & 0.876          & 28.66           & 0.860          & 32.40           & 0.933          & 32.82           & 0.930          & 32.39           & 0.916          & 30.75          & 0.903          \\
			MPRNet \cite{zamir2021multi}     & 30.27          & 0.897          & 30.41           & 0.890          & 36.40           & 0.965          & 33.64           & 0.938          & {\ul 32.91}           & 0.916          & 32.73          & 0.921          \\
			DGUNet \cite{mou2022deep}     & {\ul 30.32}    & 0.899          & {\ul 30.66}     & 0.891          & {\ul 37.42}     & 0.969          & 33.68           & 0.938          & \textbf{33.23}     & \textbf{0.920}    & {\ul 33.06}    & 0.923          \\
			KiT \cite{lee2022knn}       & 30.26          & 0.904          & 30.47           & 0.897          & 36.65           & 0.969          & {\ul 33.85}     & {\ul 0.941} & 32.81           & 0.918          & 32.81          & {\ul 0.926}    \\
			Uformer-B \cite{wang2022uformer} & 29.90          & {\ul 0.906}    & 30.31           & \textbf{0.900} & 36.86           & \textbf{0.972}    & 33.53           & 0.939    & 29.45           & 0.903          & 32.01          & 0.924          \\
			IDT \cite{xiao2022image}       & 29.69          & 0.905          & 29.95           & 0.898    & 37.01           & 0.971         & 33.38           & 0.937          & 31.38           & 0.908          & 32.28          & 0.924          \\
			Ours          & \textbf{31.69} & \textbf{0.919} & \textbf{30.90}  & {\ul 0.899}          & \textbf{37.93}  & {\ul 0.971} & \textbf{34.03}  & \textbf{0.942} & 32.11  & {\ul 0.919} & \textbf{33.33} & \textbf{0.930} \\ \hline
		\end{tabular}
	}
	\label{table1}
\end{table*}%

\subsection{Overall Framework}
Given a rainy image $I_{rain} \in \mathbb{R}^{H \times W \times 3}$, where $\mathrm{H} \times \mathrm{W}$ denotes the spatial resolution of the feature map, we utilize a $3 \times 3$ convolution to expand the feature to a higher dimensional feature space.
Following the mainstream design \cite{xiao2022image}, we stack $N_{i \in[0,1,2,3,4]}$ dual attention Transformer blocks (DATBs) in the network pipeline. 
Similar to \cite{zamir2022restormer}, each level of encoder-decoder network covers its own specific channel dimension and spatial resolution.
Besides, we add skip-connections to bridge across continuous intermediate features to ensure stable model training.
Instead of directly predicting a rain-free image $I_{derain}$, our method generates a residual image $I_{res}$ to which the input image is added to obtain: $I_{derain}=I_{rain}+I_{res}$.

In each DATB, given the input features at the ($l$-1)-th block $\mathbf{X}_{l-1}$, the encoding procedures of DATB can be formulated as
\begin{equation}
\mathbf{X}_l^{\prime}=\mathbf{X}_{l-1}+\text{DDSA}\left(\text{LN}\left(\mathbf{X}_{l-1}\right)\right),
\end{equation}
\begin{equation}
\mathbf{X}_l=\mathbf{X}_l^{\prime}+\text{SEFN}\left(\text{LN}\left(\mathbf{X}_l^{\prime}\right)\right),
\end{equation}

where $\mathbf{X}_l^{\prime}$ and $\mathbf{X}_l$ represent the outputs from the dynamic dual self-attention (DDSA) and spatial-enhanced feed-forward network (SEFN). Here, LN refers to the layer normalization.

Same with previous work \cite{zamir2022restormer}, we utilize the pixel-wise loss by the $L1$ constraint to impose supervision and optimize our model end-to-end using the following loss function:
\begin{equation}
\mathcal{L}_{pixel}=\left\|I_{derain}-I_{g t}\right\|_1,
\end{equation}
where $I_{derain}$ and $I_{g t}$ represent the output derained image and the ground-truth image, respectively.

\subsection{Dynamic Dual Self-Attention (DDSA)}
In the DATB, we develop the dynamic dual self-attention (DDSA) to replace the standard
self-attention in Transformer.
Specifically, we first encode channel-wise context by applying $1 \times 1$ convolutions followed by $3 \times 3$ depth-wise convolutions.
Given the query $Q$, key $K$, and value $V$, we generate the attention values $P$ of all pixel pairs between $Q$ and $K$:
\begin{equation}
P=\frac{Q K^{\mathrm{T}}}{\sqrt{d}},
\end{equation}
where $d = C / k$ is the head dimension and $k$ is the head number. In terms of sparse self-attention, a simple but effective masking function $\mathcal{M}(\cdot)$ is performed upon $P$ to select the top-\emph{k} largest elements from each row of the similarity matrix \cite{chen2023learning}. For other elements that are smaller than threshold, we replace them with 0. This step can further filter out noisy (irrelevant) tokens and speed up the training process, which is calculated by
\begin{equation}
\mathcal{M}(P, k)_{i j}= \begin{cases}
P_{i j} & \text { if } P_{i j} \geq \text { threshold } \\
0 & \text { if } P_{i j} <\text { threshold }
\end{cases}
\end{equation}
where threshold is a $k^{t h}$ largest value of row. Finally, the weighted sum of DSA and SSA is  matrix multiplied by $V$ to obtain the final output of DDSA, given by:
\begin{equation}
Attention=[\operatorname{softmax}(P)+\operatorname{softmax}(\mathcal{M}(P, k))]V,
\end{equation}
where $[\cdot]$ stands for weighted summation.

\begin{table*}[!t]
	\centering
	\caption{Comparison of quantitative results on real-world rainy images, lower scores indicate better image quality.}
	\resizebox{1.0\textwidth}{!}{
		\begin{tabular}{cccccccc}
			\hline
			Methods        & Rainy Image  & MSPFN \cite{jiang2020multi}    & MPRNet \cite{zamir2021multi}   & DGUNet \cite{mou2022deep}   & Uformer-B \cite{wang2022uformer} & IDT \cite{xiao2022image}      & Ours                  \\ \hline
			NIQE / BRISQUE & 5.961 / 34.147 & 4.947 / 33.027 & 4.821 / 32.116 & 4.419 / 27.654 & 4.537 / 28.619 & 4.227 / 26.237 & \textbf{4.006 / 24.773} \\ \hline
		\end{tabular}%
	}
	\label{table2}
\end{table*}

\subsection{Spatial-Enhanced Feed-forward Network (SEFN)}

Generally, Transformers also use a feed-forward network to further improve feature representation. Instead of the gated-Dconv feed-forward network (GDFN) \cite{zamir2022restormer} which only considers channel-wise information, we develop a spatial-enhanced feed-forward network (SEFN) in Transformer. In fact, rich multi-scale spatial representation has fully demonstrated its effectiveness \cite{jiang2020multi} in better removing rain. As shown in Fig. \ref{fig2}, we first employ Res2Net \cite{gao2019res2net} to encode multi-scale features at a granular level. Here, we set the scale dimension to 3. Note that we adopt $3 \times 3$ depth-wise convolutions in Res2Net for better interaction. This design complements the features with rich cross-channel and cross-spatial information. Meanwhile, the spatial attention (SA) module \cite{woo2018cbam} is inserted into SEFN to further effectively model the spatial information of features.

\section{Experiments}
\label{sec:pagestyle}

In this section, comprehensive image deraining experiments is performed on commonly used benchmark datasets to evaluate the effectiveness of the proposed method against 12 state-of-the-art image deraining baselines, \emph{i.e.}, DerainNet \cite{fu2017clearing}, SEMI \cite{wei2019semi}, DIDMDN \cite{zhang2018density}, UMRL \cite{yasarla2019uncertainty}, RESCAN \cite{li2018recurrent}, PReNet \cite{ren2019progressive}, MSPFN \cite{jiang2020multi}, MPRNet \cite{zamir2021multi}, DGUNet \cite{mou2022deep}, KiT \cite{lee2022knn}, Uformer-B \cite{wang2022uformer}, and IDT \cite{xiao2022image}.

\begin{table}[!t]
	\centering
	\caption{Ablation study analysis on the Rain100H benchmark dataset.}
	\resizebox{0.9\columnwidth}{!}{
		\begin{tabular}{cccccc}
			\hline
			Models      & (a)          & (b)         & (c)                     & (d)               & (e)   \\
			DSA         & \checkmark   &             & \checkmark              & \checkmark        & \checkmark      \\
			SSA         &              & \checkmark  & \checkmark              & \checkmark        & \checkmark     \\
			GDFN        &              &             &                         &                   & \checkmark      \\
			MSFN        & \checkmark   & \checkmark  & \checkmark              & \checkmark        &      \\
			SA          & \checkmark   & \checkmark  & \checkmark              &                   & \checkmark     \\ \hline
			PSNR        & 28.68        & 29.38       & \textbf{30.90}          & 28.96             & 29.09 \\
			SSIM        & 0.866        & 0.882       & \textbf{0.899}          & 0.881             & 0.881 \\ \hline
		\end{tabular}%
	}
	\label{table3}
\end{table}

\subsection{Datasets and Metrics}
Following \cite{li2022tao,li2023dilated}, we conduct experiments on the large-scale Rain13K dataset which contains 13, 711 paired training data. For testing sets, five common benchmarks (\emph{i.e.}, Test100 \cite{zhang2019image}, Rain100H \cite{yang2017deep}, Rain100L \cite{yang2017deep}, Test2800 \cite{fu2017removing}, and Test1200 \cite{zhang2018density}) are considered for evaluation. Here, we use peak signal-to-noise ratio (PSNR) and structural similarity index (SSIM) \cite{wang2004image}, which are two commonly used metrics to measure the deraining performance on Y channel of images in the YCbCr space. Furthermore, real-world datasets \cite{yang2020single} also considered to further evaluate the generalization performance. As ground-truth images are not available, we adopt two non-reference indicators (NIQE \cite{mittal2012making} and BRISQUE \cite{mittal2012no}).

\begin{figure*}[!t]
	\centering
	\includegraphics[width=1.0\textwidth]{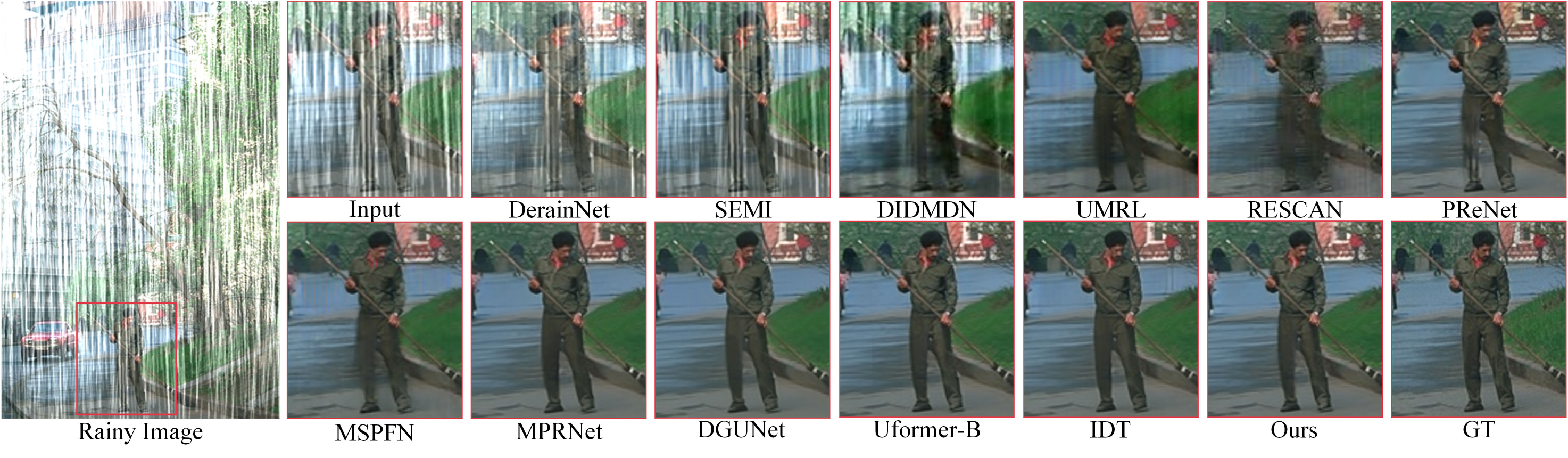} 
	\caption{Visual quality comparison of deraining images obtained by different methods on the Rain100H benchmark dataset.}
	\label{fig3}
\end{figure*}

\begin{figure*}[!t]
	\centering
	\includegraphics[width=1.0\textwidth]{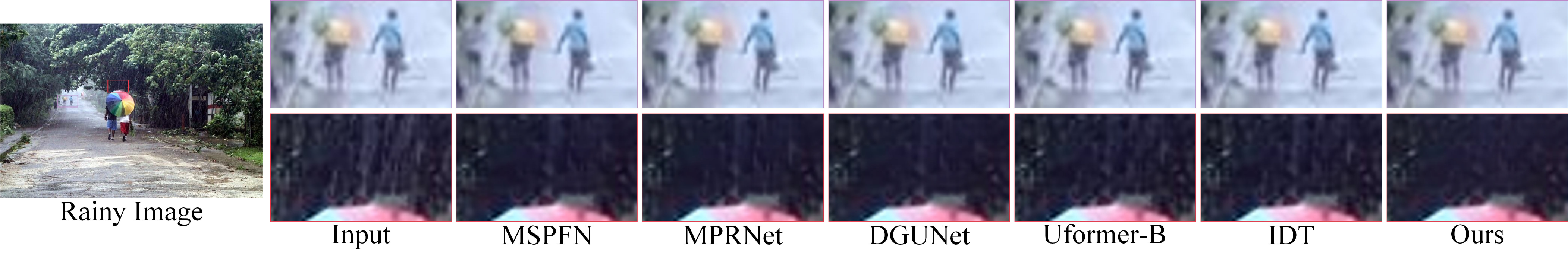} 
	\caption{Visual quality comparison of deraining images obtained by different methods on real-world rainy images.}
	\label{fig4}
\end{figure*}

\begin{figure*}[!t]
	\centering
	\includegraphics[width=1.0\textwidth]{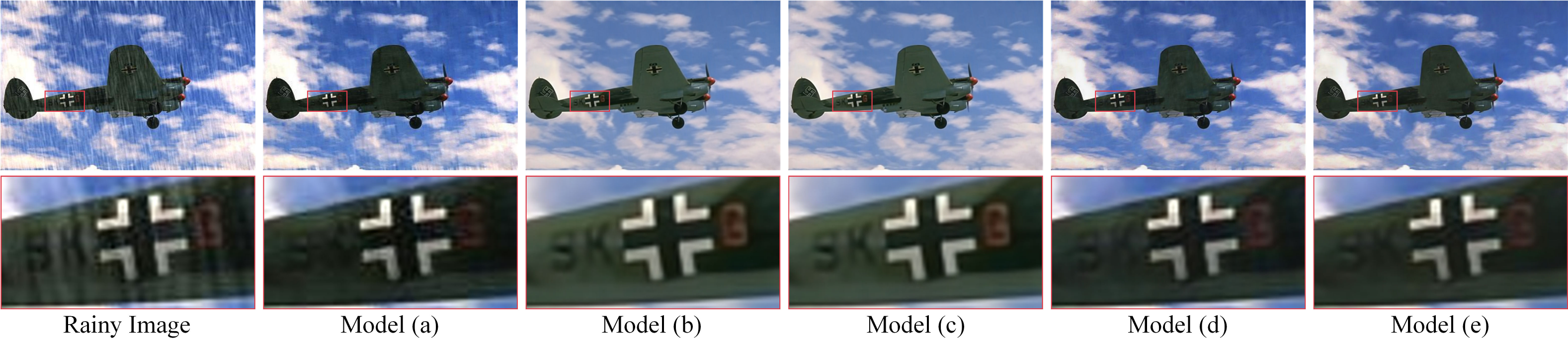} 
	\caption{Ablation qualitative comparison for different variants of our method. The models (a-e) are consistent with the settings in Table \ref{table3}.}
	\label{fig5}
\end{figure*}

\subsection{Implementation Details}
\label{sec:typestyle}

In our network settings, $\left\{N_0, N_1, N_2, N_3, N_4\right\}$ are set to $\{2,4,6,6,8\}$, and the number of attention heads for 5 DATB of the same level is set to $\{1,1,2,4,8\}$. The head number $k$ is set to 0.7. All experiments are implemented by PyTorch and trained on 4 NVIDIA GeForce RTX 3090 GPUs. Each dataset is trained with total 300K iterations, using AdamW optimizer with a batch size of 16 and a patch size of 128. The initial learning rate is fixed as $1 \times 10^{-4}$ for 92K iterations, and then adopt the cosine annealing to adjust the learning rate progressively. Moreover, we augment the training data by flipping (horizontal and vertical) to enhance the robustness of our network.  

\subsection{Comparison with State-of-the-arts}
{\flushleft \textbf{Synthetic Datasets}.}
Table \ref{table1} reports the average PSNR and SSIM computed on the different benchmarks. Note that as Uformer-B and IDT are not trained on the Rain13K \cite{jiang2020multi}, we retrain their models to ensure fair comparisons. Our method achieves the highest average scores in terms of PSNR and SSIM. Fig. \ref{fig3} shows the visual comparsions on the Rain100H dataset. We can clearly observe that compared to other deraining approaches, our method accomplish the high-quality restoration in rain removal and detail preservation, thanks to the performance advantages of our developed framework.

{\flushleft \textbf{Real-world Datasets}.}
We further compare our method with other competing approaches on real-world rainy images. From the quantitative results in Table \ref{table2}, our network obtains the lower NIQE and BRISQUE values, which means high-quality deraining results with clearer content and better perceptual quality. From Fig. \ref{fig4}, it can be observed that the proposed method significantly competes others in removing most rain streaks, showing great potential in real-world situations.

\subsection{Ablation Studies}
We study the main component impacts on the final
deraining performance. Here, we mainly consider the following variants of our method: (1) DAS, SSA or DDSA; (2) GDFN or MSFN; (3) w or w/o SA. Table \ref{table3} lists the corresponding quantitative evaluation on the Rain100H. We note that our model (c) performs better than the other possible configurations, which remarkably reveals that each individual component has a positive impact on performance improvement. As presented in Fig. \ref{fig5}, our method can restore clearer rain-free background.

\section{Conclusion}
\label{sec:majhead}

In this work, we have proposed an effective image deraining Transformer with dynamic dual self-attention. We combine the developed sparse self-attention with the dense self-attention in standard Transformer, and a spatial-enhanced feed-forward network, in order to better facilitate rain removal and help image restoration. The quantitative and qualitative experiments on a variety of rain streak benchmarks have been conducted to demonstrate the superiority of our proposed method. In the future, we will explore the potential of this framework in other image restoration tasks.

%
%
%
%



\bibliographystyle{IEEEbib}
\bibliography{refs}

\begin{thebibliography}{10}

\bibitem{yu2022towards}
Yi~Yu, Wenhan Yang, Yap-Peng Tan, and Alex~C Kot,
\newblock ``Towards robust rain removal against adversarial attacks: A
  comprehensive benchmark analysis and beyond,''
\newblock in {\em IEEE CVPR}, 2022, pp. 6013--6022.

\bibitem{fan2023learning}
Zhentao Fan, Xianhao Wu, Xiang Chen, and Yufeng Li,
\newblock ``Learning to see in nighttime driving scenes with inter-frequency
  priors,''
\newblock in {\em IEEE CVPRW}, 2023, pp. 4217--4224.

\bibitem{kang2011automatic}
Li-Wei Kang, Chia-Wen Lin, and Yu-Hsiang Fu,
\newblock ``Automatic single-image-based rain streaks removal via image
  decomposition,''
\newblock {\em IEEE TIP}, vol. 21, no. 4, pp. 1742--1755, 2011.

\bibitem{li2016rain}
Yu~Li, Robby~T Tan, Xiaojie Guo, Jiangbo Lu, and Michael~S Brown,
\newblock ``Rain streak removal using layer priors,''
\newblock in {\em IEEE CVPR}, 2016, pp. 2736--2744.

\bibitem{fu2017clearing}
Xueyang Fu, Jiabin Huang, Xinghao Ding, Yinghao Liao, and John Paisley,
\newblock ``Clearing the skies: A deep network architecture for single-image
  rain removal,''
\newblock {\em IEEE TIP}, vol. 26, no. 6, pp. 2944--2956, 2017.

\bibitem{chen2022unpaired}
Xiang Chen, Jinshan Pan, Kui Jiang, Yufeng Li, Yufeng Huang, Caihua Kong,
  Longgang Dai, and Zhentao Fan,
\newblock ``Unpaired deep image deraining using dual contrastive learning,''
\newblock in {\em IEEE CVPR}, 2022, pp. 2017--2026.

\bibitem{yang2020single}
Wenhan Yang, Robby~T Tan, Shiqi Wang, Yuming Fang, and Jiaying Liu,
\newblock ``Single image deraining: From model-based to data-driven and
  beyond,''
\newblock {\em IEEE TPAMI}, vol. 43, no. 11, pp. 4059--4077, 2020.

\bibitem{li2022tao}
Yufeng Li, Zhentao Fan, Jiyang Lu, and Xiang Chen,
\newblock ``Tao-net: Task-adaptive operation network for image restoration and
  enhancement,''
\newblock {\em IEEE SPL}, vol. 29, pp. 2198--2202, 2022.

\bibitem{zamir2022restormer}
Syed~Waqas Zamir, Aditya Arora, Salman Khan, Munawar Hayat, Fahad~Shahbaz Khan,
  and Ming-Hsuan Yang,
\newblock ``Restormer: Efficient transformer for high-resolution image
  restoration,''
\newblock in {\em IEEE CVPR}, 2022, pp. 5728--5739.

\bibitem{xiao2022image}
Jie Xiao, Xueyang Fu, Aiping Liu, Feng Wu, and Zheng-Jun Zha,
\newblock ``Image de-raining transformer,''
\newblock {\em IEEE TPAMI}, 2022.

\bibitem{dosovitskiy2020image}
Alexey Dosovitskiy, Lucas Beyer, Alexander Kolesnikov, Dirk Weissenborn,
  Xiaohua Zhai, Thomas Unterthiner, Mostafa Dehghani, Matthias Minderer, Georg
  Heigold, Sylvain Gelly, et~al.,
\newblock ``An image is worth 16x16 words: Transformers for image recognition
  at scale,''
\newblock {\em arXiv}, 2020.

\bibitem{chen2023hybrid}
Xiang Chen, Jinshan Pan, Jiyang Lu, Zhentao Fan, and Hao Li,
\newblock ``Hybrid cnn-transformer feature fusion for single image deraining,''
\newblock in {\em AAAI}, 2023, vol.~37, pp. 378--386.

\bibitem{chen2022unpaired2}
Xiang Chen, Zhentao Fan, Pengpeng Li, Longgang Dai, Caihua Kong, Zhuoran Zheng,
  Yufeng Huang, and Yufeng Li,
\newblock ``Unpaired deep image dehazing using contrastive disentanglement
  learning,''
\newblock in {\em ECCV}. Springer, 2022, pp. 632--648.

\bibitem{wang2022uformer}
Zhendong Wang, Xiaodong Cun, Jianmin Bao, Wengang Zhou, Jianzhuang Liu, and
  Houqiang Li,
\newblock ``Uformer: A general u-shaped transformer for image restoration,''
\newblock in {\em IEEE CVPR}, 2022, pp. 17683--17693.

\bibitem{zhang2019image}
He~Zhang, Vishwanath Sindagi, and Vishal~M Patel,
\newblock ``Image de-raining using a conditional generative adversarial
  network,''
\newblock {\em IEEE TCSVT}, vol. 30, no. 11, pp. 3943--3956, 2019.

\bibitem{yang2017deep}
Wenhan Yang, Robby~T Tan, Jiashi Feng, Jiaying Liu, Zongming Guo, and Shuicheng
  Yan,
\newblock ``Deep joint rain detection and removal from a single image,''
\newblock in {\em IEEE CVPR}, 2017, pp. 1357--1366.

\bibitem{fu2017removing}
Xueyang Fu, Jiabin Huang, Delu Zeng, Yue Huang, Xinghao Ding, and John Paisley,
\newblock ``Removing rain from single images via a deep detail network,''
\newblock in {\em IEEE CVPR}, 2017, pp. 3855--3863.

\bibitem{zhang2018density}
He~Zhang and Vishal~M Patel,
\newblock ``Density-aware single image de-raining using a multi-stream dense
  network,''
\newblock in {\em IEEE CVPR}, 2018, pp. 695--704.

\bibitem{wei2019semi}
Wei Wei, Deyu Meng, Qian Zhao, Zongben Xu, and Ying Wu,
\newblock ``Semi-supervised transfer learning for image rain removal,''
\newblock in {\em IEEE CVPR}, 2019, pp. 3877--3886.

\bibitem{yasarla2019uncertainty}
Rajeev Yasarla and Vishal~M Patel,
\newblock ``Uncertainty guided multi-scale residual learning-using a cycle
  spinning cnn for single image de-raining,''
\newblock in {\em IEEE CVPR}, 2019, pp. 8405--8414.

\bibitem{li2018recurrent}
Xia Li, Jianlong Wu, Zhouchen Lin, Hong Liu, and Hongbin Zha,
\newblock ``Recurrent squeeze-and-excitation context aggregation net for single
  image deraining,''
\newblock in {\em ECCV}, 2018, pp. 254--269.

\bibitem{ren2019progressive}
Dongwei Ren, Wangmeng Zuo, Qinghua Hu, Pengfei Zhu, and Deyu Meng,
\newblock ``Progressive image deraining networks: A better and simpler
  baseline,''
\newblock in {\em IEEE CVPR}, 2019, pp. 3937--3946.

\bibitem{jiang2020multi}
Kui Jiang, Zhongyuan Wang, Peng Yi, Chen Chen, Baojin Huang, Yimin Luo, Jiayi
  Ma, and Junjun Jiang,
\newblock ``Multi-scale progressive fusion network for single image
  deraining,''
\newblock in {\em IEEE CVPR}, 2020, pp. 8346--8355.

\bibitem{zamir2021multi}
Syed~Waqas Zamir, Aditya Arora, Salman Khan, Munawar Hayat, Fahad~Shahbaz Khan,
  Ming-Hsuan Yang, and Ling Shao,
\newblock ``Multi-stage progressive image restoration,''
\newblock in {\em IEEE CVPR}, 2021, pp. 14821--14831.

\bibitem{mou2022deep}
Chong Mou, Qian Wang, and Jian Zhang,
\newblock ``Deep generalized unfolding networks for image restoration,''
\newblock in {\em IEEE CVPR}, 2022, pp. 17399--17410.

\bibitem{lee2022knn}
Hunsang Lee, Hyesong Choi, Kwanghoon Sohn, and Dongbo Min,
\newblock ``Knn local attention for image restoration,''
\newblock in {\em IEEE CVPR}, 2022, pp. 2139--2149.

\bibitem{chen2023learning}
Xiang Chen, Hao Li, Mingqiang Li, and Jinshan Pan,
\newblock ``Learning a sparse transformer network for effective image
  deraining,''
\newblock in {\em IEEE CVPR}, 2023, pp. 5896--5905.

\bibitem{gao2019res2net}
Shang-Hua Gao, Ming-Ming Cheng, Kai Zhao, Xin-Yu Zhang, Ming-Hsuan Yang, and
  Philip Torr,
\newblock ``Res2net: A new multi-scale backbone architecture,''
\newblock {\em IEEE TPAMI}, vol. 43, no. 2, pp. 652--662, 2019.

\bibitem{woo2018cbam}
Sanghyun Woo, Jongchan Park, Joon-Young Lee, and In~So Kweon,
\newblock ``Cbam: Convolutional block attention module,''
\newblock in {\em ECCV}, 2018, pp. 3--19.

\bibitem{li2023dilated}
Yufeng Li, Jiyang Lu, Hongming Chen, Xianhao Wu, and Xiang Chen,
\newblock ``Dilated convolutional transformer for high-quality image
  deraining,''
\newblock in {\em IEEE CVPRW}, 2023, pp. 4198--4206.

\bibitem{wang2004image}
Zhou Wang, Alan~C Bovik, Hamid~R Sheikh, and Eero~P Simoncelli,
\newblock ``Image quality assessment: from error visibility to structural
  similarity,''
\newblock {\em IEEE TIP}, vol. 13, no. 4, pp. 600--612, 2004.

\bibitem{mittal2012making}
Anish Mittal, Rajiv Soundararajan, and Alan~C Bovik,
\newblock ``Making a “completely blind” image quality analyzer,''
\newblock {\em IEEE SPL}, vol. 20, no. 3, pp. 209--212, 2012.

\bibitem{mittal2012no}
Anish Mittal, Anush~Krishna Moorthy, and Alan~Conrad Bovik,
\newblock ``No-reference image quality assessment in the spatial domain,''
\newblock {\em IEEE TIP}, vol. 21, no. 12, pp. 4695--4708, 2012.

\end{thebibliography}

\end{document}